\documentclass[sigconf]{acmart}
\AtBeginDocument{%
  \providecommand\BibTeX{{%
    \normalfont B\kern-0.5em{\scshape i\kern-0.25em b}\kern-0.8em\TeX}}}

\usepackage[utf8]{inputenc} % allow utf-8 input
\usepackage[T1]{fontenc}    % use 8-bit T1 fonts
\usepackage{hyperref}       % hyperlinks
\usepackage{url}            % simple URL typesetting
\usepackage{booktabs}       % professional-quality tables
\usepackage{amsfonts}       % blackboard math symbols
\usepackage{nicefrac}       % compact symbols for 1/2, etc.
\usepackage{microtype}      % microtypography
\usepackage{lipsum}
\usepackage{graphicx}
\usepackage{amsmath}
\usepackage{amsthm}

\theoremstyle{definition}

\usepackage[export]{adjustbox}
\usepackage[justification=centering]{subfig}
\usepackage{float}
\usepackage{xcolor}
\usepackage{multicol}

\begin{document}

\title{On the Bias-Variance Characteristics of LIME and SHAP in High Sparsity Movie Recommendation Explanation Tasks}


\author{Claudia V. Roberts}
\authornote{Both authors contributed equally to this research.}
\email{claudiar@princeton.edu}
\affiliation{%
  \institution{Princeton University}
  \country{USA}
}

\author{Ehtsham Elahi}
\authornotemark[1]
\email{eelahi@netflix.com}
\affiliation{%
  \institution{Netflix, Inc.}
  \country{USA}
}

\author{Ashok Chandrashekar}
\email{ashok.chandrashekar@warnermedia.com}
\affiliation{%
  \institution{Warner Media}
  \country{USA}
}

%\author{
%  Claudia Roberts \thanks{The two authors have equal contribution}\footnotemark[1]\\\
%%  Department of Computer Science\\
 %% Princeton University\\
 %% Princeton, NJ 08544 \\
 %% \texttt{claudiar@princeton.edu} \\
 %% \And
 %% Ehtsham Elahi \footnotemark[1]\\
 %% Product ML Research\\
 %% Netflix, Inc.\\
 %% Los Gatos, CA 95032 \\
 %% \texttt{eelahi@netflix.com} \\
  %% examples of more authors
 %% \And
 %%Ashok Chandrashekar \\
 %% Product ML Research\\
 %% Netflix, Inc.\\
 %% Los Gatos, CA 95032 \\
 %% \texttt{achandrashekar@netflix.com}} \\


\begin{abstract}
We evaluate two popular local explainability techniques, LIME and SHAP, on a movie recommendation task. We discover that the two methods behave very differently depending on the sparsity of the data set. LIME does better than SHAP in dense segments of the data set and SHAP does better in sparse segments. We trace this difference to the differing bias-variance characteristics of the underlying estimators of LIME and SHAP. We find that SHAP exhibits lower variance in sparse segments of the data compared to LIME. We attribute this lower variance to the completeness constraint property inherent in SHAP and missing in LIME. This constraint acts as a regularizer and therefore increases the bias of the SHAP estimator but decreases its variance, leading to a favorable bias-variance trade-off especially in high sparsity data settings. With this insight, we introduce the same constraint into LIME and formulate a novel local explainabilty framework called Completeness-Constrained LIME (CLIMB) that is superior to LIME and much faster than SHAP.
\end{abstract}

\maketitle

\section{Introduction}
Recommendation systems mediate our various online interactions on a daily basis by limiting and influencing our possible choices. Recommender system use cases include product recommendations, search engines, social media browsing, music and video streaming, online advertising, news dissemination, job candidate matching, and real estate recommendations. The recommendation system problem setting is a high sparsity problem both from the perspective of the user and the system making the recommendations. From the perspective of the user, the user only has prior information on a tiny subset of the total number of items at her disposable. From the perspective of the system, the system has very little interaction data for the vast majority of the available items. This makes the recommendation setting an important and challenging problem domain. 

In the recommender domain, explanations can be an integral part of the user product experience and depending on the recommendation task, critical to the task description itself. Explanatory models provide explanations for why the underlying recommendation system model made the item selection, item position ranking, or point prediction estimate that it did. In this paper, we focus on local explanations, that is, explanations for a single prediction instance. Two popular, general purpose explanation frameworks whose aim is to faithfully explain the local predictions of machine learning models are Local Interpretable Model-agnostic Explanations (LIME) and SHapley Additive exPlanations (SHAP).  LIME is very easy to use, computationally fast, and works on tabular data, images, and text \cite{molnar2019}. While SHAP is computationally much slower than LIME depending on the underlying prediction model, it has some important theoretical guarantees such as guaranteeing the fair distribution of the prediction across the features \cite{shap, molnar2019}. 

The first research question we sought to answer was how do SHAP and LIME perform in the high sparsity recommendation system setting. We adapted LIME and SHAP to the task of explaining movie recommendations and evaluated the explanations using the delta-rank metric (described in Section \ref{sec:experiments}). We observed that while SHAP outperforms LIME on aggregate, the two methods behave very differently depending on the sparsity of the data. LIME does better than SHAP in dense segments of the data set, and conversely, SHAP outperforms LIME in the sparse regions of the data set. We performed a bias-variance analysis and traced this difference in performance to the differing bias and variance characteristics of the underlying estimators of LIME and SHAP. We show that SHAP exhibits lower variance and higher bias compared to LIME and we postulate that this is the reason why SHAP outperforms LIME in high sparsity data settings where the bias-variance trade-off is especially favorable. 

We hypothesize that the reason for SHAP's lower variance is due to Shapley values satisfying the efficiency property or what other papers call the completeness axiom \cite{integ-grad}, the conservation property \cite{lrp}, or summation-to-delta property \cite{deep-grad} (for the duration of this paper we will refer to this property as the completeness constraint). We argue that this completeness constraint acts as a regularizer and therefore increases the bias and decreases the variance of the SHAP estimator. With these collective insights supported by our analysis, we introduce this constraint into LIME; we call this new local explainability technique Completeness-Constrained LIME (CLIMB). Our experiments show that CLIMB indeed lowers the variance of the LIME estimator and improves its performance in sparse data settings. CLIMB allows users to enjoy some of the theoretical guarantees of SHAP and maintain the off-the-shelf ease of LIME whilst being computationally faster than LIME and improving performance in high sparsity data settings, commonplace in recommendation tasks.

Our contributions are summarized as follows:
\begin{itemize}
    \item A comparison between SHAP and LIME in a movie recommendation setting, specifically analyzing their performance in sparse and dense regions of a publicly available data set 
    \item A bias-variance analysis of SHAP and LIME in the sparse and dense data regions in a movie recommendation setting
    \item Formulation of a new model-agnostic, faithful, local explanation method called CLIMB that includes one of the powerful properties of SHAP while being as fast as LIME and maintaining some of the desirable qualities of LIME
    \item Analysis connecting bias and variance to the completeness constraint 
\end{itemize}

\section{Background} \label{sec:background}
When determining what items to present to a user, these systems necessarily pare down the complete set of possible items from the millions to a small handful. The recommendation system problem setting is a high sparsity problem where the recommending system has very little interaction data between all the available users and all the available items \cite{Kumar_sparse, Zellou_sparse, Sahu_sparse, AHMADIAN_sparse, chen_sparse}. Recommendation systems can also suffer from the long-tail phenomenon---there is an outsized amount of user interaction data for a tiny subset of the available item set and an extremely large number of items which effectively have no interaction data \cite{mmds}. Further contributing to the high sparsity nature of online recommenders is the highly dynamic and in some cases transitory nature of the data. Users and product items are constantly coming and going, whether physically or in terms of relevancy, and user tastes are ever evolving. 

An important aspect of recommendation systems is their corresponding explanatory models. This tight coupling of recommendation system models and explanation models is unique to the recommendation system setting. In the computer vision domain, explanations might come in the form of a visual saliency map that indicates the specific pixels that most contributed to the prediction of ``cat'' in an image classification task, for example. In the natural language processing task of sentiment analysis, an explanation model might highlight the particular words in a social media comment that most contributed to the comment being flagged as inappropriate by the model. In both of these cases, explanations serve largely as sanity checks to ensure that the learned mathematical model is picking up on the right features. Explanations in these artificial intelligence domains help build confidence that the trained machine learning models are doing the right thing for the right reasons and not picking up on spurious features. 

The goals for providing explanations in recommendation systems and for sometimes explicitly exposing them to the user as a product feature are numerous and as follows: transparency, validation, trust building, persuasion, effectiveness, efficiency, satisfaction, communicating relevancy, comprehensibility, educating \cite{Tintarev07}. Previous studies have shown that accompanying recommendations with their explanations lead to higher user acceptance of recommendations though care must be taken because poorly designed explanations can be less performant than the base case of no explanations at all \cite{Herlocker2000, persuasive, tintarev2012}. In the computer vision example of image classification (and other mundane automation tasks), if the user of the system is 100\% confident that the system is correct 100\% of the time then there is no need for explanations---a cat is a cat, is a cat yesterday, today, and tomorrow. In the highly dynamic world of item recommendation where there are competing incentives, explanations can be used to surprise and delight users as well as build trust amongst multiple stakeholders. Today, a user might hate horror movies but tomorrow, that same user might be delighted to be recommended a particular horror movie because it is top trending in the country and he wants to be part of that moment, part of the cultural zeitgeist. 

Evaluating the explanations of a single model prediction instance is separated into two components 1) faithfulness of the explanation 2) ease of human understanding \cite{lime, Hoeve2018FaithfullyER, DoshiVelez2017TowardsAR}. An explanatory model is said to be locally faithful if the predictive behavior of the explanatory model in the vicinity of the single instance of interest is consistent with the predictive behavior of the underlying recommender model in the same vicinity. An explanatory model is said to be intelligible or interpretable if the explanation for a single recommendation instance is readily understood by a human. Evaluating the ease of human understanding of a local explanation is highly subjective and task dependent and not the focus of this paper. Studying the faithfulness of an explanation model is important because a low-fidelity explanation, an explanation that does not closely approximate the behavior of the underlying model, means that the explanation model is not accurately or honestly describing the underlying recommender model's decision making process \cite{Herman2017ThePA, Lipton18}. 

\section{Related Work} \label{sec:relatedwork}
Two of the most popular model-agnostic local explanation methods are LIME and SHAP. LIME learns a separate interpretable model trained on a new data set of random permutations of the original data instance we are seeking to explain \cite{lime}. SHAP explains the prediction of an individual data instance by computing Shapley values \cite{shap}. Shapley values is a game theoretic technique that estimates the contribution of each feature to the prediction also by perturbing the original input data instance \cite{shapleyvals}. Previous work comparing SHAP and LIME focuses on evaluating these explanation methods based on their stability or reproducibility, that is, their ability to return consistent explanations over numerous runs on the same input \cite{slime, trusty, visani_stable, Man2020TheBW, Mythreyi_stab_fid}. Other work evaluating explanation frameworks assesses their local fidelity or faithfulness to the original underlying model \cite{fidel_mess, du_liu, exact_consistent, robustness_paper, Mythreyi_stab_fid}. Additionally, a common paradigm when evaluating and comparing SHAP, LIME, and other explanation methods is to introduce a new evaluation metric and evaluate the explanations against this metric, e.g. effectiveness, efficiency, necessity, sufficiency, XAI Test, feature importance similarity, feature importance consistency, impact score, impact coverage \cite{Yanou, mothilal2021towards, jesus, many_faces, Lin2019DoER, interp_health, more_health}. Most recently, researchers evaluated the robustness of LIME and SHAP and found them to be vulnerable to adversarial attacks where the explanatory models can be manipulated to hide potentially harmful biases in the original model \cite{fooling_lime_shap, domen_robust}. 

As we highlighted in Section \ref{sec:background}, the recommender setting requires domain specific consideration given the unique technical challenges it poses and the unique and various needs it has for explanations. To the best of our knowledge, we are the first to evaluate SHAP and LIME based on their performance in different data sparsity settings. More concretely, to the best of our knowledge, we are the first to evaluate these explanation models based on how they perform when explaining a recommendation for a data instance with very little historical interaction data versus when explaining a recommendation for a data instance with plentiful historical interaction data. We are also the first to connect this difference in data-sparsity-dependent performance to the differing bias-variance characteristics of SHAP and LIME and subsequently, the completeness constraint that is inherent in SHAP but missing in LIME. We then go on to prove this hypothesis by formulating a novel explanation method called Completeness-Constrained LIME (CLIMB) that indeed improves the performance of LIME in sparse data settings.

\section{Preliminaries} \label{sec:prelims}
In this section, we lay down the mathematical foundation and build up the theoretical scaffolding necessary for understanding our ensuing contributions. 
\subsection{LIME}
Local Interpretable Model-agnostic Explanations (LIME) is a framework for training a secondadry interpretable model, or surrogate model, to explain the individual predictions coming from any opaque classifier \cite{lime}. The LIME algorithm for training a surrogate model works as following. First, select some data instance $x \in\mathbb{R}^d$ for which you want an explanation, i.e. you want an explanation for why an opaque recommender model $f$ predicted that user feature vector $x$ would play/not play a movie with probability $f(x)$. LIME requires that in order for the explanation to be understandable to humans, the data should be transformed into an interpretable representation such as a binary vector $x' \in{\{0,1\}}^{d'}$ denoting the presence/absence of interpretable components, e.g. user watched/did not watch movie A in the past. Next, generate a new data set $Z$ of perturbed samples $z' \in{\{0,1\}}^{d'}$ by drawing nonzero elements of $x'$ uniformly at random. Now that we have a new set of data instances $Z$ in the neighborhood of $x'$, we need labels for them. To obtain the labels needed for our new explanatory model, we transform the perturbed samples $z'$ back into their original representation $z \in\mathbb{R}^d$ and interrogate the opaque model for each instance $f(z)$. Because we randomly generated the perturbed samples $z'$ we would like to capture the fact that some samples $z$ might be closer or farther to the original data instance of interest $x$ and thus should be weighted accordingly. This weighting scheme is captured by the proximity measure  $\pi_x(z)$, which measures the proximity between an instance $x$ to $z$. 

Finally, using this new weighted data set $Z$ and ground truth labels generated by obtaining $f(Z)$ we train a new model $g \in G$ where G is a class of interpretable models such as decision trees, linear models, etc. This new model $g$ is our interpretable, explanatory surrogate model $\xi(x)$ for explaining $f(x)$: 
\begin{equation}
\xi(x) = \arg\min_{g \in G} L(f,g,\pi_x) + \Omega(g)
\end{equation}
$L$ is any loss function of your choice which measures how unfaithful $g$ is at approximating the behavior of $f$ in the local neighborhood of $x$. We want to minimize this loss function so that the behavior of $g$ mimics the behavior of $f$ as closely as possible in the locality defined by $\pi_x$. $\Omega(g)$ is a complexity term of the model---we want this to be low, e.g. we prefer fewer features in the case of linear models. In the original LIME paper, the authors use the square loss function $L$ with $\ell_2$ penalty. Typically, $g(z')$ is chosen to be a linear function i.e. $g(z') = \Phi^Tz' + \phi_0$ which makes the above a weighted linear regression problem to solve for $\Phi$ and intercept $\phi_0$. 
\begin{equation}
L(f,\Phi, \phi_0,\pi_x) = \sum_{z,z'\in Z} \pi_x(z)(f(z)-(\phi_0 + \Phi^Tz'))^2
\end{equation}
 
Some of the advantages of LIME include 1) off-the-shelf easy to use implementation available 2) relatively fast computationally 3) works with tabular data, text, and images 4) opaque model can change without needing to change the explanation model implementation \cite{molnar2019}. Some of the previously reported disadvantages of LIME include 1) many hyperparameters to set whose choices heavily influence the resulting explanation and leads to many scientific degrees of freedom (perturbation sampling strategy, neighborhood definition, selection of $g$) 2) instability of explanation output as mentioned in Section \ref{sec:relatedwork} 3) no theoretical guarantees that would help the LIME explanation of a prediction hold up in court \cite{limitations, molnar2019}. To the best of our knowledge, we are the first to shine a light on LIME's decreased performance in high sparsity data regions as well as highlight its comparatively good performance in dense data regions as compared to SHAP in a recommender setting. 

\subsection{SHAP}
Like LIME, SHapley Additive exPlanations is an attribution method, that is, a method that describes the prediction of a single data instance as the sum of the effects each feature had on the prediction \cite{molnar2019}. Shapley values is an explanation framework that explains the prediction of an individual data instance by computing Shapley values \cite{shap, shapleyvals, molnar2019}. We choose the model-agnostic Kernel Shap formulation (denoted as SHAP in the rest of the paper) which describes the local explanation as a weighted linear regression similar to LIME as shown in equation [1] with $g(z') = \Phi^Tz' + \phi_0$. The regression loss function and the weights are given by:
\begin{equation}
\begin{aligned}
    L(f,\Phi, \phi_0,\pi_x) &= \sum_{z,z'\in Z} \pi_x(z)(f(z)- (\phi_0 + \Phi^Tz'))^2 \\
    \pi_x'(z') &= \frac{d' - 1}{(d'\ choose\ |z'|)|z'|(d' - |z'|)}
\end{aligned}
\end{equation}

where $d'$ is the dimensionality of $x'$ and $|z'|$ is the number of non-zero elements in $z'$. In contrast to LIME, generation of the data set $Z$ is very different in SHAP. In SHAP, $Z$ is defined as the power set of all non-zero indices in $x'$. Hence, $Z$ has a size of $2^{d'}$ if we exhaustively enumerate all possible subsets. (Typical software implementations do allow putting an upper limit on the number of samples in $Z$). Therefore, one of the computational complexities of SHAP is generating this data set $Z$. Another (minor) difference from LIME is that the regularization parameter in Shapley values regression $\Omega(g) = 0$.

As shown in \cite{shap}, this choice of weighting function $\pi_x'(z') = \infty$ when $|z'| \in \{0, d'\}$. This means for $z' = 0 \implies z=\emptyset$ (a null or baseline feature vector) and $\phi_0 = f(\emptyset)$ and $z' = x' \implies \phi_0 + \sum_{i=1}^{d'}\phi_i = f(x)$ (since all zeroes can be dropped from $x'$ as missing/zero features have no contribution, therefore $x'$ is simply a vector of all ones and $\Phi^Tx' = \sum_{i=1}^{d'}\phi_i$). This is the so-called completeness constraint. SHAP calls this the local accuracy property \cite{shap}, Shapley values calls it the efficiency property \cite{shapleyvals} and yet other papers call it the completeness axiom \cite{integ-grad}, the conservation property \cite{lrp}, or the summation-to-delta property \cite{deep-grad}. For the duration of this paper we will refer to this property as the completeness constraint. We discuss the implications of this constraint in detail in the next section. 

Some of the advantages of SHAP include 1) the prediction of a single instance is fairly distributed among the feature values 2) game theoretic guarantees afforded to it by Shapley values  \cite{molnar2019}. Some of the previously reported disadvantages of SHAP include 1) slow computation due to high computational complexity 2) like LIME, SHAP is also vulnerable to adversarial attacks and has issues with explanation instability \cite{molnar2019}. To the best of our knowledge, we are the first to show SHAP's decreased performance in dense regions of the data set  its superior performance in sparse regions of the data set as compared to LIME in a recommender setting. Furthermore, we are the first to trace this difference in performance to SHAP's lower variance in high sparsity data settings, which we show is a result from SHAP satisfying the completeness constraint. 

\subsection{Completeness Constraint} \label{sec:complete}
The completeness constraint $f(x) = f(\emptyset) + \sum_{i=1}^{d'}\phi_i$ has two immediate computational implications:
\begin{itemize}
    \item The intercept of the regression function is set to $f(\emptyset)$ and is no longer a free parameter and thus, does not need to be estimated.
    \item $\Phi$ has $d'-1$ degrees of freedom. For example, the $d'$-th element of $\Phi$ can be written as  $\phi_{d'} = f(x) - f(b) - \sum_{i=1}^{d'-1}\phi_i$.
\end{itemize}
We would like to comment that $z' = 0$ does not need to correspond to a literal zero/empty feature vector and can be chosen to be any feature vector $b$ as long as $f(x) \neq f(b)$. We do however use a zero feature vector as the null/baseline feature vector in this paper (more details to come in Section \ref{sec:experiments}) and use $\emptyset$ to denote it for the remainder of this paper. 
Qualitatively, the completeness property has a number of implications:
\begin{itemize}
    \item Under the completeness constraint, the $\Phi$ is said to have a fair attribution of feature importance as it captures the contribution of each feature in the underlying model's prediction at data instance $x$. LIME is simply a best-fit line and the learned linear function may not be equal to $f$ at the data instance $x$.
    \item If the data instance $x$ and the baseline $b$ is different in only one feature, then the differing feature is given a non-zero attribution under the completeness constraint (since $f(x) \neq f(b) \forall x \neq b$). To see how we might end up with zero attribution for features without this constraint we reference the example given in \cite{integ-grad}. Consider a function $f(x) = 1 - \text{ReLU}(1-x))$ and say we want a local explanation at $x = 2$. This function changes from $0$ to $1$ at $x=1$ and after that it becomes flat. A local explainability method like LIME may result in a regression line with $0$ slope due to the local flatness of the function. But choosing $b=0$ where $f(0) = 0$ would force Shapley values to learn a regression model with a non-zero slope. Therefore, for highly non-linear recommendation models that may have many such flat regions, the completeness constraint helps generate accurate explanations in such ``zero-gradient'' sub-regions in the feature landscape.
\end{itemize}

\section{CLIMB: Completeness-Constrained LIME} \label{climb}
\subsection{Preliminary Experimental Results Motivating CLIMB}
In this section, we briefly summarize our initial experimental findings on a movie recommender explanation task that served as the catalyst for the resultant body of research. Full implementation details along with a detailed description of the evaluation metric we used can be found in Section \ref{sec:experiments}.

Knowing how important explanations can be to the product experience of recommendation systems and knowing that these systems suffer greatly from having either no previous interaction data (the cold-start problem) or very little historical interaction data (in comparison to the available item set), we wanted to evaluate how well SHAP and LIME perform in varying data sparsity settings. In our first experiment, shown in Figure \ref{fig:allmethodcomp}, we iteratively removed the $top-k$ most important features from the data instance of interest $x$. We observed that as we increased the number of features that we removed from $x$, the gap in performance between SHAP and LIME widened, with SHAP outperforming LIME. In a second preliminary experiment, shown in Figure \ref{fig:allmethodcomp_sparse}, we divided our movie recommendation data set into eight equal sized groups based on sparsity, i.e. based on the amount of interaction data each data instance had. We observed that SHAP significantly outperformed LIME in the sparsest groups and that LIME outperformed SHAP in the densest groups. This interesting reversal of performance based on the sparsity of the data has been observed many times in machine learning research \cite{bishop_bias_var} and has been found to be closely related to the bias-variance characteristics of models.

Both SHAP and LIME attempt to predict the behavior of an underlying model in the neighborhood of the given data instance $x$. Their ability to provide the correct explanations is therefore tied to their generalization ability in the local neighborhood around $x$. We can decompose the generalization capability in terms of their bias and variance. To be precise, since both SHAP and LIME are regression models, their generalization error can be measured in terms of the following mean-squared error.

\begin{equation}
\begin{aligned}
    \textit{MSE}(x; \Phi) &= \mathbb{E}[(f(x) - \hat{\Phi}(x))^2]\\
    &= (f(x) - \mathbb{E}[\hat{\Phi}(x)])^2 + \mathbb{E}[(\hat{\Phi}(x) - \mathbb{E}(\hat{\Phi}(x))^2]\\
    &= \textit{Bias}^2 + \textit{Variance}
\end{aligned}
\end{equation}
where $\hat{\Phi}$ is an estimator of $\Phi$. (Note: if the underlying model is non-stochastic, there is no residual error term).

Bootstrapping is one straightforward way to compute the bias and variance of any model. For the explanation models, the bootstrapping procedure proceeds by generating $P$ local perturbations of $x$ by randomly zeroing out features. For the $p$-th perturbed vector, we solve the explanation model to get $\hat{\Phi}_p$. So the empirical average $\mathbb{E}[\hat{\Phi}(x)] \approx \frac{\sum_{p = 1}^{P} \hat{\Phi}_p(x)}{P}$ can be plugged-in to estimate the bias and variance in the above equation. Note that this bias and variance is meant to capture the behavior of the explanation model in the neighborhood of $x$.

With these analysis tools, we conducted a bias-variance analysis of SHAP and LIME (results shown in Figure \ref{fig:bias_var_sparse}) on the same eight sparsity groups from the previous evaluation experiment. We observed that in all segments of the dataset, SHAP exhibited higher bias and lower variance. In the sparsest segments, there was a big variance reduction with a small increase in bias resulting in a favorable bias-variance trade-off. This favorable bias-variance trade-off leads to SHAP improving upon LIME significantly in the sparsest regions of the dataset. In the denser regions, there is a small variance reduction with a large increase in the bias resulting in SHAP's poor performance compared to LIME. This analysis provides strong evidence that the behavior of SHAP and LIME with respect to data sparsity can be easily explained in terms of their bias-variance characteristics. We hypothesize that this bias-variance difference arises due to the completeness constraint (present in SHAP and missing in LIME) which we discuss in the next section.

\subsection{The Bias-Variance and Completeness Constraint Connection}
Our preliminary findings showed that SHAP and LIME perform differently depending on the density or sparsity of the data instance whose prediction we seek an explanation for. We showed that this difference is statistically significant. After conducting a bias-variance analysis of SHAP and LIME, we observed that SHAP exhibits lower variance than LIME in high sparsity data regions. As we stated in Section \ref{sec:background}, high sparsity data regions are commonplace in recommendation systems and thus, it is important that these explanation frameworks perform well in high sparsity settings. We posit that the completeness constraint property, inherent in SHAP and missing in LIME, is an important reason for why SHAP outperforms LIME in sparse data settings. In this section, we reason how the completeness constraint is tied to the observable bias-variance characteristics of SHAP, thus foreshadowing the motivation behind our novel completeness-constrained explanation model. 

Given that SHAP enjoys the same game theoretic grounding as Shapley values, including the completeness constraint, we asked ourselves the following research question, ``How is the completeness constraint connected to the bias-variance behavior exhibited by SHAP in sparse data regions?'' The completeness constraint was originally motivated by the desire for attribution methods to fairly distribute the prediction among the features and served as a solution to the gradient saturation problem mentioned in Section \ref{sec:complete}. However, given our interest in explanations for recommender systems, we take an entirely different approach to analyzing its role in the performance of SHAP vs. LIME in sparse data settings. 

Since the completeness constraint limits the flexibility of the explanation model, by eliminating both the intercept and one degree of freedom from $\Phi$, we argue that it plays the same role as a regularizer. In other words, the limited flexibility prevents the explanation model's regression function from fully fitting the behavior of the underlying model in the neighborhood of the data instance $x$, thus resulting in increased bias. But this reduced flexibility would also reduce variance of the explanation model. As long as this bias-variance trade-off is favorable (for example in sparse settings), we expect to see improved accuracy in predicting the behavior of $f$ from explanation models with the completeness constraint. Studying the bias-variance trade-off of the completeness constraint is a novel approach and forms the basis of our work. 

\subsection{Formulation of CLIMB}
As mentioned in Section \ref{sec:prelims}, LIME has highly desirable qualities such as off-the-shelf ease of use that makes it an attractive choice over the computationally slower but theoretically more sound SHAP. We propose introducing the completeness constraint into LIME to take advantage of the favorable bias-variance characteristics of SHAP. Additionally, adding this constraint into LIME would provide the fair attribution property found in SHAP and help protect against generating erroneous/zero explanations in locally flat sub-regions. We now introduce our straightforward formulation of Completeness-Constrained LIME (CLIMB).

We set up CLIMB identically to LIME. We have the data instance $x \in \mathbb{R}^d$ and its interpretable binary representation $x' \in{\{0,1\}}^{d'}$, a new data set $Z$ comprised of perturbed data samples $z'$ ($z$ in the original feature space) and their corresponding labels $f(z)$, and the proximity weighting function $\pi_x(z)$, all identical to LIME. In order to introduce the completeness constraint into LIME, we borrow the concept of a baseline feature vector $b \in R^d$ from SHAP. Like SHAP, the choice of $b$ is problem dependent. We explain our choice of $b$ for the recommendation model we use in Section \ref{sec:experiments}. 

CLIMB is the solution to the following constraint least squares problem,
\begin{equation}
\begin{aligned}
\min_{\Phi} &\sum_{z,z'\in Z} \pi_x(z)(f(z) - (f(x) + \Phi^T(z' - x')))^2\\
&\textrm{s.t.}\ \Phi^Tx' = f(x) - f(b)\\
\end{aligned}
\end{equation}

Note that the intercept of the above regression function is $f(b)$ like SHAP. The solution $\Phi \in R^{d'}$ is a vector of coefficients and is interpreted in the same way as the solution for LIME and SHAP. Fortunately, we do not have to solve the above constraint optimization directly since that would make CLIMB computationally slower than LIME. The completeness constraint is a linear constraint, and we can eliminate the constraint by the following substitution. First, note that $\Phi^Tx' = \sum_{j=1}^{d'} \phi_j$. Therefore, we can substitute out $\phi_{d'} = f(x_0) - f(b) - \sum_{j = 1}^{d' - 1} \phi_j$ in the above equation. Let $c=f(b) + x'_{d'} (f(x) - f(b))$ and $r(z') = (z'_{1:d'-1} - z'_{d'})$, then the first $d'-1$ components of $\Phi$ (denoted below as $\Phi_{1:d'-1}$) are obtained by the following unconstrained least squares minimization
\begin{equation}
    \min_{\Phi_{1:d'-1}} \sum_{z, z' \in Z} \pi_x(z) (f(z) - ( c + r(z')^T\Phi_{1:d'-1}))^2 
\end{equation}
The last component of $\Phi$ (denoted as $\Phi_{d'}$ above) is obtained by back substituting in the linear constraint. This way of solving for $\Phi$ results in an algorithm that should be as fast as LIME as the problem dimension is reduced to having one less degree of freedom compared to LIME and there is no intercept to estimate. 

\section{Experiments} \label{sec:experiments}
\subsection{Experimental Setup}
\subsubsection{Model} We use a Multinomial Variational Autoencoder (Mult-VAE) \cite{multi-vae} trained on the MovieLens 20M data set \cite{movielens} as the recommendation model whose predictions we want to explain. MovieLens is a data set of users that interacted with movies on the MovieLens website. For the Mult-VAE model, each user is a represented as a bag-of-words of movies that they interacted with. Therefore, the feature vector $x_u$ for a user $u$ can be represented as k-hot binary vector of size 20,108 (total number of movies in the data set) with $1's$ for the interacted movies and $0's$ for the rest. For any user represented as this k-hot encoded vector, Mult-VAE model can score the entire collection of 20,108 movies. Typically, these scores are then used to rank the entire collection of movies (in descending order) to generate personalized recommendations/rankings.  

\subsubsection{Data Preparation} For our local explanability experiments, we use the validation split of 10,000 users outside of the training set. For each validation user $u$, we generated the personalized ranking from the Mult-VAE model and use the top-ranked movie $t_u$ for local explanability. Therefore the data instance $x_u$ is the k-hot vector and $f_{t_u}(x)$ is the score of the Mult-VAE model for the top-ranked movie. Note that the corresponding interpretable version of $x$ is a vector $x'$ of size $d'$ of all ones where $d'$ is the number of non-zero entries in $x$. From this vector $x$, the data set $Z$ can be generated by sampling the non-zero indices and therefore are binary vectors of size $d'$. This data set generation strategy is same for LIME and CLIMB whereas it is different for SHAP, as described in Section \ref{sec:prelims}. We do control for the number of samples in $Z$ and keep it fixed to 5,000 for the three explanation methods. Our evaluation metric (described next) requires a ranking of non-zero movies in $x$, therefore we turn off the $\ell_1$ penalty in SHAP and any feature selection heuristic in LIME so that we may get explanation coefficients $\Phi$ for all non-zero movies in the data instance $x_u$. We keep the rest of the parameters fixed to their default values. For both SHAP and CLIMB, the choice of baseline is a zero feature vector meaning a null user without any interaction history. The Mult-VAE model outputs an unpersonalized score for each movie when this zero feature vector is used as input. The unpersonalized score is proportional to the number of non-zero interactions for each movie in the training data(typically called the training data popularities of movies in the recommendation models literature).

\subsubsection{Evaluation Metric} We quantitatively evaluate the explanation methods using the delta-prediction metric (also seen in other papers as the ``change in log-odds'' \cite{Sippy2020DataSA, deep-grad, shap, cxplain}) and adapt it to the recommendation task and call it the delta-rank metric. Given a ranking of non-zero movies in $x_u$ according to the explanation model coefficient $\phi_i, i=1,...,d'$, for each validation user $u$, take the \textit{top-k input movies} according to the explanation model coefficients and remove them from $x_u$. This gives a modified data instance ${x_u}_m$ which is the same as $x_u$ except for the missing movies that we removed. Compute the output ranking from the Mult-VAE model with ${x_u}_m$ as the input. Calculate the difference in the rank of the movie $t_u$, which was the top ranked movie earlier. The idea is that if the movies that were removed from $x_u$ were really important for the Mult-VAE to rank $t_u$ at the top, we should expect to see a big drop in the ranking of $t_u$. We remove a large number of movies (for example up-to $30$) by taking a few of them at a time (for example $6$ at a time) and plot the change in the rank (or delta-rank) as we remove each batch of $6$ movies. We expect the delta-rank to be negative if important features are removed, and the magnitude of the drop to be proportional to the importance of features removed (therefore lower the better). We compute summary statistics of this delta-rank metric for all validation users. 

\begin{figure}
  \centering
  \includegraphics[width=0.8\columnwidth]{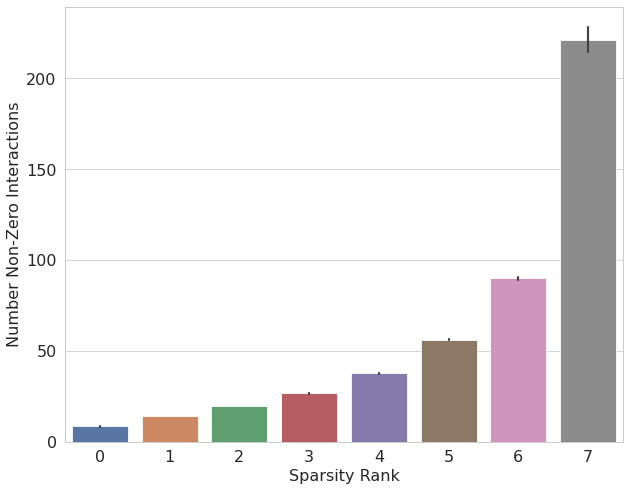}
  \caption{Number of non-zero movie interactions in each sparsity segment}
  \label{fig:sparse_stats}
\end{figure}

Since we are interested in comparing the bias-variance and delta-rank performance of SHAP, LIME and CLIMB for different sparsity settings, we partition the 10,000 validation users in eight equal sized buckets according to the number of non-zero movie interactions in feature vector $x_u$. In the results below, we label the data set segment with the highest sparsity as Sparsity Rank = $0$ and the lowest sparsity segment as Sparsity Rank = $7$. Figure-\ref{fig:sparse_stats} describes the sparsity characteristics of each segment. 

Our results can be fully reproduced using the the Jupyter Notebooks found in the supplementary materials. 

\subsection{Results}
\subsubsection{Delta-rank Comparison Among LIME, SHAP, CLIMB}
\begin{figure}
  \centering
  \includegraphics[width=0.8\columnwidth]{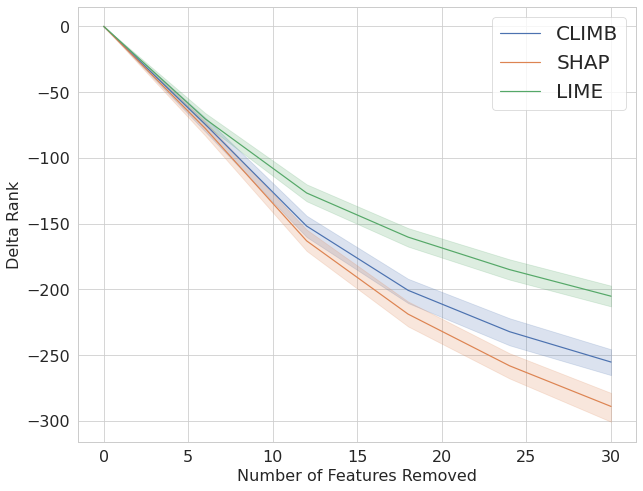}
  \includegraphics[width=0.8\columnwidth]{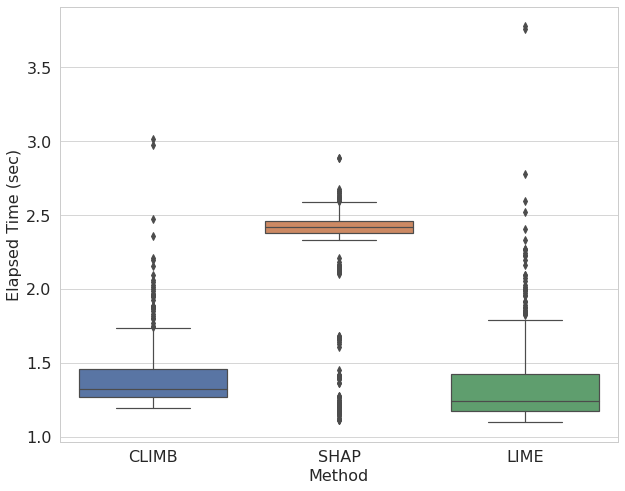}
  \caption{Comparing CLIMB, SHAP and LIME according to delta-rank and computational speed}
  \label{fig:allmethodcomp}
\end{figure}

\begin{figure}
  \centering
  %\fbox{\rule[-.5cm]{4cm}{4cm} \rule[-.5cm]{4cm}{0cm}}
  %\includegraphics[width=85mm]{fd_shap_lime.png}
    %\includegraphics[width=1.0\columnwidth]{kdd_results/sparse_comp_new_1.png}
    \includegraphics[width=1.0\columnwidth]{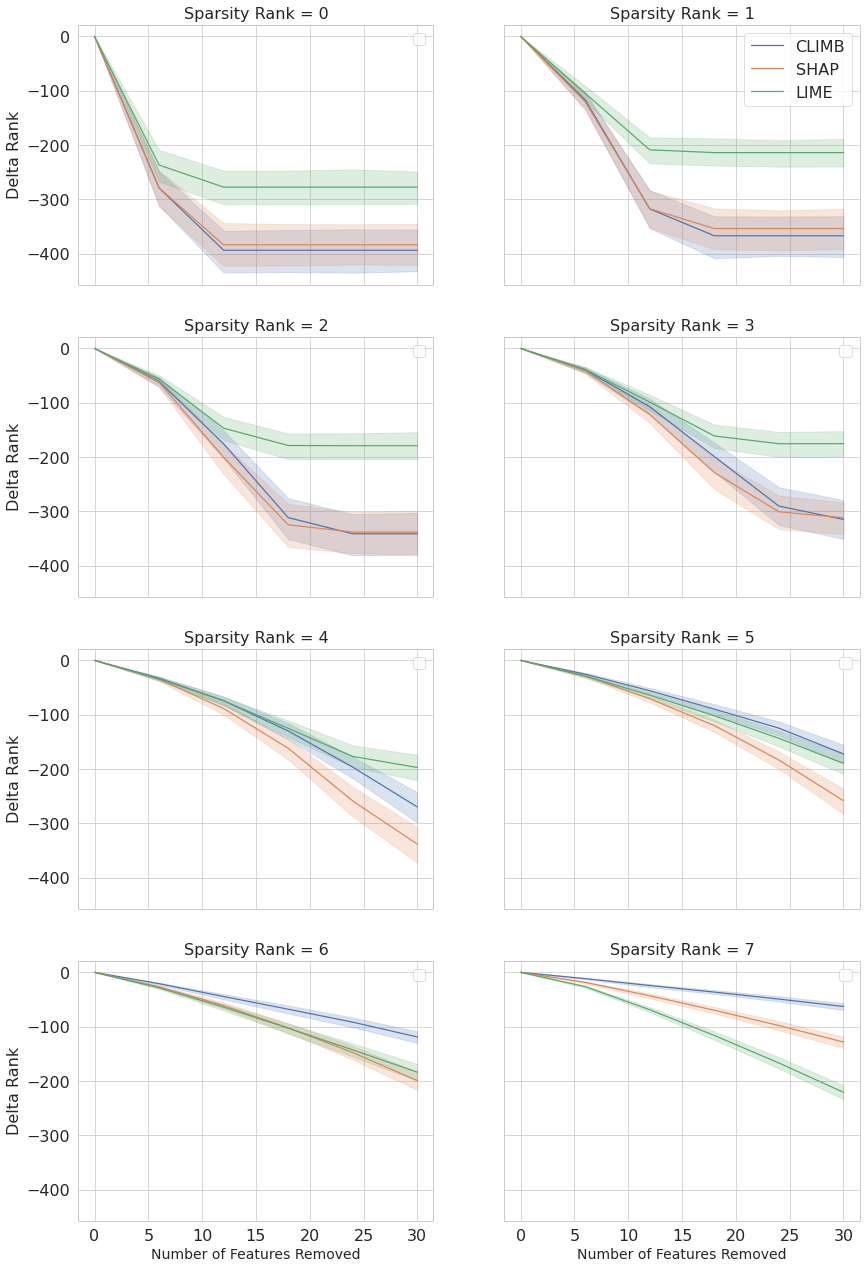}
  \caption{Comparing CLIMB, SHAP and LIME in decreasing order of sparsity. Sparsity Rank = 0 is the data set segment with highest sparsity and Sparsity Rank = 7 has the least sparsity}
  \label{fig:allmethodcomp_sparse}
\end{figure}

As shown in Figure-\ref{fig:allmethodcomp}, both CLIMB and SHAP outperform LIME significantly whereas the difference between CLIMB and SHAP is insignificant up to top-20 features. This validates our hypothesis that introducing the completeness constraint into LIME does indeed result in improved local explanability.  We also compare the three methods according to sparsity using the eight segments described above (Figure-\ref{fig:allmethodcomp_sparse}). We see the expected outcome---the overall delta-rank improvements come from the sparse segments of the data set where CLIMB and SHAP outperform LIME. We attribute this improvement to an overall favorable bias-variance trade-off especially in the sparse segments of the MovieLens data set.

\subsubsection{Computational Analysis}
As mentioned earlier, integrating the completeness constraint into LIME results in an estimation problem of lower complexity and can be solved as fast as LIME. The second figure in Figure-\ref{fig:allmethodcomp} shows this result.
\subsubsection{Bias-Variance Analysis of LIME, SHAP, CLIMB}
\begin{figure}
  \centering
  %\fbox{\rule[-.5cm]{4cm}{4cm} \rule[-.5cm]{4cm}{0cm}}
  %\includegraphics[width=85mm]{fd_shap_lime.png}
  \includegraphics[width=0.80\columnwidth]{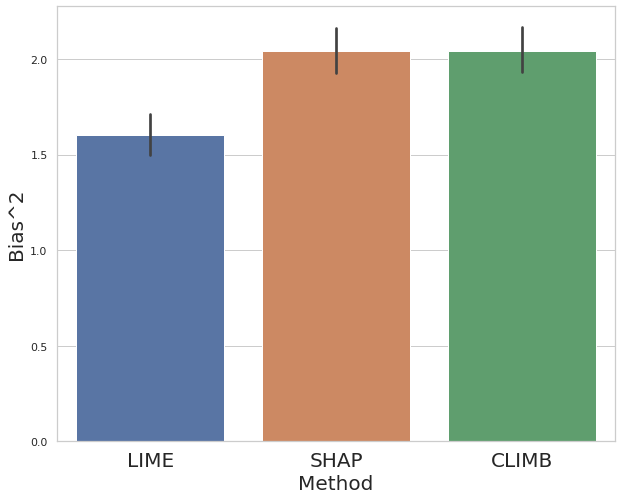}
    \includegraphics[width=0.80\columnwidth]{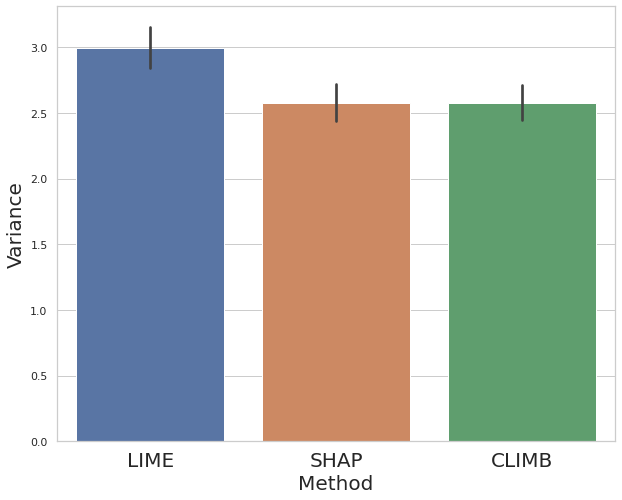}
  \caption{Comparing the overall Bias and Variance of CLIMB, SHAP and LIME}
  \label{fig:bias_var_overall}
\end{figure}

\begin{figure}
  \centering
  %\fbox{\rule[-.5cm]{4cm}{4cm} \rule[-.5cm]{4cm}{0cm}}
  %\includegraphics[width=85mm]{fd_shap_lime.png}
  \includegraphics[width=0.80\columnwidth]{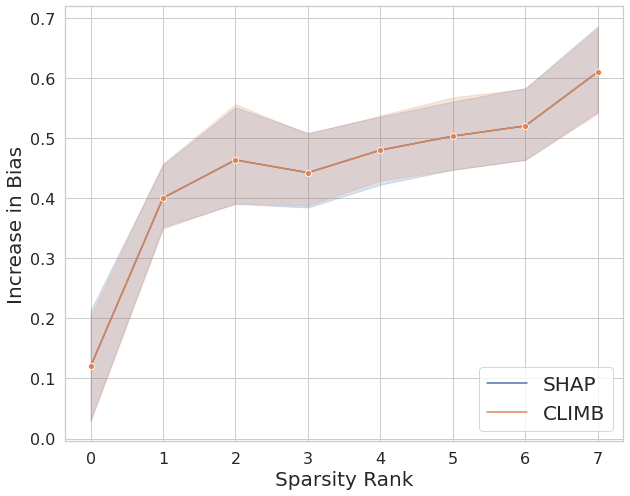}
    \includegraphics[width=0.80\columnwidth]{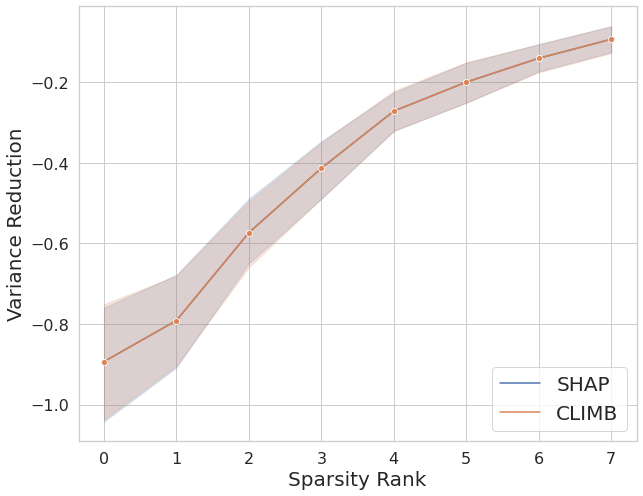}
  \caption{Comparing the Change in Bias and Variance of CLIMB and SHAP relative to LIME with decreasing levels of sparsity}
  \label{fig:bias_var_sparse}
\end{figure}

We use a validation set of size 1,000 for bias-variance computation (down from 10,000 to keep the computation time in check) and we solve LIME, SHAP and CLIMB estimation problems for $50$ bootstrapped perturbation of each validation example. Figure-\ref{fig:bias_var_overall} shows that indeed CLIMB and SHAP exhibit higher bias and lower variance as we hypothesized in the earlier section. Moreover, Figure-\ref{fig:bias_var_sparse} shows that the variance reduction (compared to LIME) is directly proportional to the sparsity whereas increase in bias (compared to LIME) is inversely proportional to the sparsity. These results show that we get the best bias-variance trade-off in the most sparse segments of the data set. Our results also show the role the completeness constraint plays as a regularization technique, therefore significantly improving the performance of LIME by incorporating completeness constraint in it in the sparse segments of the MovieLens dataset.  
\subsubsection{Qualitatively Examining Local Explanations}
We find examples where the delta-rank metric for CLIMB is far better than LIME to build an intuition for how  improvements in delta-rank affect the outward quality of the resulting explanations. ``Star Wars : Empire Strikes Back'' and ``Harry Potter and The Goblet of Fire'' are two such examples selected from the sparse region of the MovieLens data set. Looking at the explanations visually, the results for both CLIMB and SHAP are identical and qualitatively much better than LIME (we highlight the explanations in red that subjectively seem to make little sense). Looking at these explanations and noting the improvements in the delta-rank metric, we conclude that these explanations not only visually make sense but are in-agreement with the underlying model. We note that the metric or a visual examination alone will not allow us to make this claim. We also include one example from the dense region of the data set, ``Star Trek: The Wrath of Khan'', where the delta-rank metric for LIME is superior to CLIMB and SHAP. CLIMB seems to include a number of seemingly unrelated movies in its explanations. According to our analysis, the bias-variance trade-off due to the completeness constraint is unfavorable in the dense regions and this is reflected in the subjective quality of the explanations as well. 
\begin{figure}
  \centering
  %\fbox{\rule[-.5cm]{4cm}{4cm} \rule[-.5cm]{4cm}{0cm}}
  %\includegraphics[width=85mm]{fd_shap_lime.png}
  \includegraphics[width=1\columnwidth]{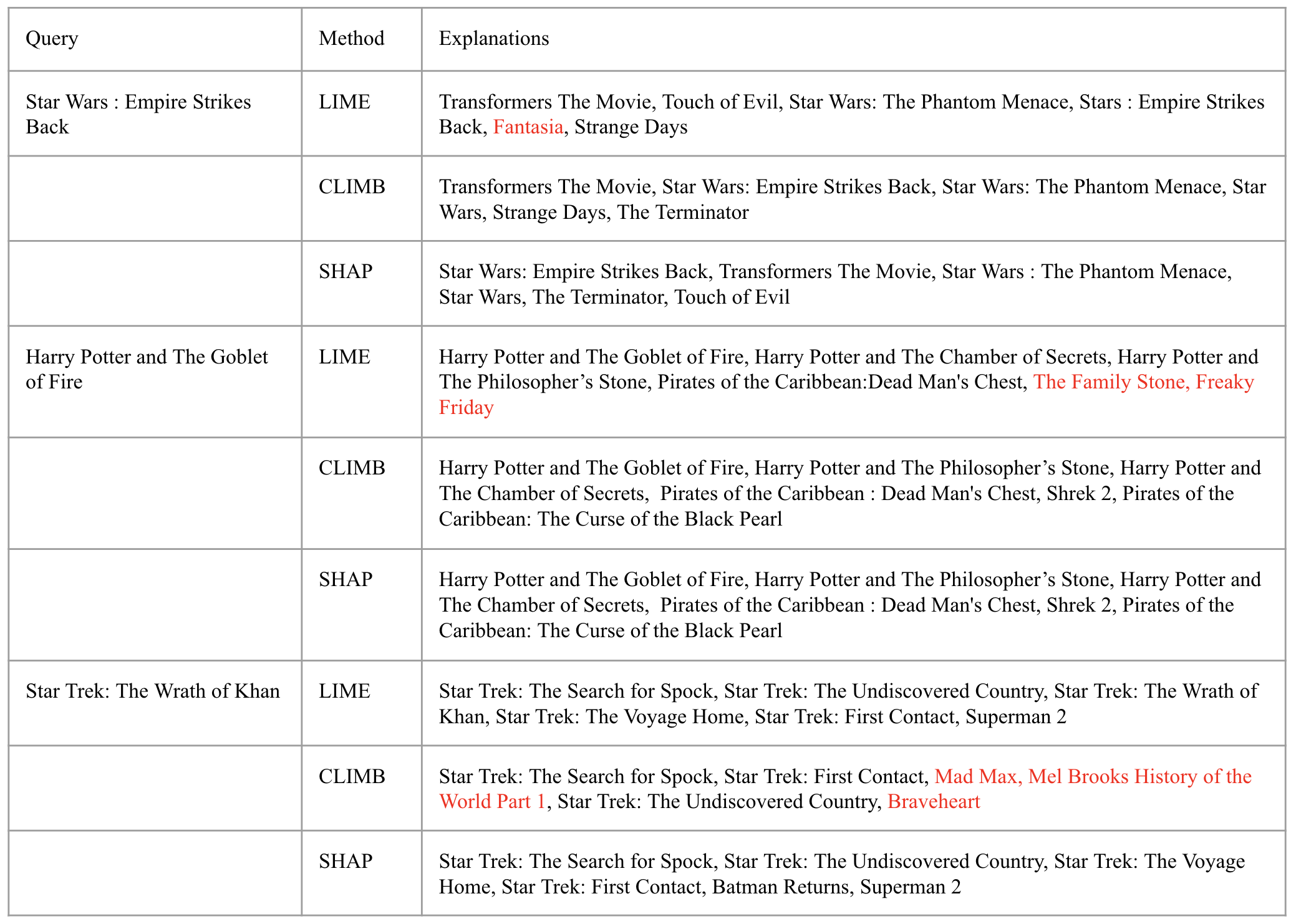}
  \caption{Comparing explanations generated for two sparse and one dense queries}
  \label{fig:explanations}
\end{figure}

\section {Conclusion}
In this paper, we 1) provided motivation for why explanations for recommender systems require special consideration, 2) showed the shortcomings LIME, a popular, easy to use explanation method, had in addressing the needs of recommender systems, which often operate in high sparsity data settings, 3) traced the root of the issue to an important property that is found in another popular but slower explanation method, SHAP, 4) incorporated this property into LIME to create a novel explanation framework called CLIMB, and finally, 5) showed that CLIMB is superior to LIME in high sparsity data settings, is as fast as LIME (much faster than SHAP), and is as easy to use as LIME.

\section* {Acknowledgments}
Thank you to Arvind Narayanan for enabling this body of research and for helpful discussions. This research was supported by the National Defense Science and Engineering Graduate Fellowship and in part by the National Science Foundation under Award CHS-1704444. 

\bibliographystyle{unsrt}  
\bibliography{coffee_recsys}  %%% Remove comment to use the external .bib file (using bibtex).

\begin{thebibliography}{10}

\bibitem{molnar2019}
Christoph Molnar.
\newblock {\em Interpretable Machine Learning}.
\newblock 2019.
\newblock \url{https://christophm.github.io/interpretable-ml-book/}.

\bibitem{shap}
Scott~M. Lundberg and Su-In Lee.
\newblock A unified approach to interpreting model predictions.
\newblock In {\em Proceedings of the 31st International Conference on Neural
  Information Processing Systems}, NIPS'17, page 4768–4777, Red Hook, NY,
  USA, 2017. Curran Associates Inc.

\bibitem{integ-grad}
Mukund Sundararajan, Ankur Taly, and Qiqi Yan.
\newblock Axiomatic attribution for deep networks.
\newblock In {\em Proceedings of the 34th International Conference on Machine
  Learning - Volume 70}, ICML'17, page 3319–3328. JMLR.org, 2017.

\bibitem{lrp}
Alexander Binder, Sebastian Bach, Gregoire Montavon, Klaus-Robert M{\"u}ller,
  and Wojciech Samek.
\newblock Layer-wise relevance propagation for deep neural network
  architectures.
\newblock In Kuinam~J. Kim and Nikolai Joukov, editors, {\em Information
  Science and Applications (ICISA) 2016}, pages 913--922, Singapore, 2016.
  Springer Singapore.

\bibitem{deep-grad}
Avanti Shrikumar, Peyton Greenside, and Anshul Kundaje.
\newblock Learning important features through propagating activation
  differences.
\newblock In {\em Proceedings of the 34th International Conference on Machine
  Learning - Volume 70}, ICML'17, page 3145–3153. JMLR.org, 2017.

\bibitem{Kumar_sparse}
Akshi Kumar and Abhilasha Sharma.
\newblock Alleviating sparsity and scalability issues in collaborative
  filtering based recommender systems.
\newblock In Suresh~Chandra Satapathy, Siba~K. Udgata, and Bhabendra~Narayan
  Biswal, editors, {\em Proceedings of the International Conference on
  Frontiers of Intelligent Computing: Theory and Applications (FICTA)}, pages
  103--112, Berlin, Heidelberg, 2013. Springer Berlin Heidelberg.

\bibitem{Zellou_sparse}
Nouhaila Idrissi and Zellou Ahmed.
\newblock A systematic literature review of sparsity issues in recommender
  systems.
\newblock {\em Social Network Analysis and Mining}, 10, 12 2020.

\bibitem{Sahu_sparse}
Ashish Sahu and Pragya Dwivedi.
\newblock User profile as a bridge in cross-domain recommender systems.
\newblock {\em Applied Intelligence}, 01 2019.

\bibitem{AHMADIAN_sparse}
Sajad Ahmadian, Nima Joorabloo, Mahdi Jalili, and Milad Ahmadian.
\newblock Alleviating data sparsity problem in time-aware recommender systems
  using a reliable rating profile enrichment approach.
\newblock {\em Expert Systems with Applications}, 187:115849, 2022.

\bibitem{chen_sparse}
Yibo Chen, Chanle Wu, Ming Xie, and Xiaojun Guo.
\newblock Solving the sparsity problem in recommender systems using association
  retrieval.
\newblock {\em JCP}, 6:1896--1902, 08 2011.

\bibitem{mmds}
Jure Leskovec, Anand Rajaraman, and Jeffrey~David Ullman.
\newblock {\em Mining of Massive Datasets}.
\newblock Cambridge University Press, USA, 2nd edition, 2014.

\bibitem{Tintarev07}
Nava Tintarev and Judith Masthoff.
\newblock A survey of explanations in recommender systems.
\newblock In G.~Uchyigit, editor, {\em Data Engineering Workshop}, pages
  801--810. IEEE Computer Society, December 2007.
\newblock IEEE 23rd International Conference on Data Engineering (ICDE 2007) ;
  Conference date: 16-04-2007 Through 20-04-2007.

\bibitem{Herlocker2000}
Jonathan~L. Herlocker, Joseph~A. Konstan, and John Riedl.
\newblock Explaining collaborative filtering recommendations.
\newblock In {\em Proceedings of the 2000 ACM Conference on Computer Supported
  Cooperative Work}, CSCW '00, page 241–250, New York, NY, USA, 2000.
  Association for Computing Machinery.

\bibitem{persuasive}
Sofia Gkika and George Lekakos.
\newblock The persuasive role of explanations in recommender systems.
\newblock {\em CEUR Workshop Proceedings}, 1153:59--68, 01 2014.

\bibitem{tintarev2012}
Nava Tintarev and Judith Masthoff.
\newblock Evaluating the effectiveness of explanations for recommender systems.
\newblock {\em User Modeling and User-Adapted Interaction}, 22, 10 2012.

\bibitem{lime}
Marco~Tulio Ribeiro, Sameer Singh, and Carlos Guestrin.
\newblock "why should i trust you?": Explaining the predictions of any
  classifier.
\newblock In {\em Proceedings of the 22nd ACM SIGKDD International Conference
  on Knowledge Discovery and Data Mining}, KDD '16, page 1135–1144, New York,
  NY, USA, 2016. Association for Computing Machinery.

\bibitem{Hoeve2018FaithfullyER}
Maartje ter Hoeve, Anne Schuth, Daan Odijk, and M.~de~Rijke.
\newblock Faithfully explaining rankings in a news recommender system.
\newblock {\em ArXiv}, abs/1805.05447, 2018.

\bibitem{DoshiVelez2017TowardsAR}
Finale Doshi-Velez and Been Kim.
\newblock Towards a rigorous science of interpretable machine learning.
\newblock {\em arXiv: Machine Learning}, 2017.

\bibitem{Herman2017ThePA}
Bernease Herman.
\newblock The promise and peril of human evaluation for model interpretability.
\newblock {\em ArXiv}, abs/1711.07414, 2017.

\bibitem{Lipton18}
Zachary~C. Lipton.
\newblock The mythos of model interpretability.
\newblock {\em Commun. ACM}, 61(10):36–43, sep 2018.

\bibitem{shapleyvals}
Lloyd~S. Shapley.
\newblock {\em A Value for N-Person Games}.
\newblock RAND Corporation, Santa Monica, CA, 1952.

\bibitem{slime}
Zhengze Zhou, Giles Hooker, and Fei Wang.
\newblock {\em S-LIME: Stabilized-LIME for Model Explanation}, page
  2429–2438.
\newblock Association for Computing Machinery, New York, NY, USA, 2021.

\bibitem{trusty}
Rob Geada, Tommaso Teofili, Rui Vieira, Rebecca Whitworth, and Daniele Zonca.
\newblock Trustyai explainability toolkit.
\newblock {\em CoRR}, abs/2104.12717, 2021.

\bibitem{visani_stable}
Giorgio Visani, Enrico Bagli, Federico Chesani, Alessandro Poluzzi, and Davide
  Capuzzo.
\newblock Statistical stability indices for {LIME:} obtaining reliable
  explanations for machine learning models.
\newblock {\em CoRR}, abs/2001.11757, 2020.

\bibitem{Man2020TheBW}
Xin Man and Ernest~P. Chan.
\newblock The best way to select features? comparing mda, lime, and shap.
\newblock 2020.

\bibitem{Mythreyi_stab_fid}
Mythreyi {Velmurugan}, Chun {Ouyang}, Catarina {Moreira}, and Renuka
  {Sindhgatta}.
\newblock {Evaluating Explainable Methods for Predictive Process Analytics: A
  Functionally-Grounded Approach}.
\newblock {\em arXiv e-prints}, page arXiv:2012.04218, December 2020.

\bibitem{fidel_mess}
Andreas Messalas, Christos Makris, and Yannis Kanellopoulos.
\newblock Model-agnostic interpretability with shapley values.
\newblock 07 2019.

\bibitem{du_liu}
Mengnan Du, Ninghao Liu, Fan Yang, Shuiwang Ji, and Xia Hu.
\newblock On attribution of recurrent neural network predictions via additive
  decomposition.
\newblock In {\em The World Wide Web Conference}, WWW '19, page 383–393, New
  York, NY, USA, 2019. Association for Computing Machinery.

\bibitem{exact_consistent}
Lingyang Chu, Xia Hu, Juhua Hu, Lanjun Wang, and Jian Pei.
\newblock Exact and consistent interpretation for piecewise linear neural
  networks: A closed form solution.
\newblock In {\em Proceedings of the 24th ACM SIGKDD International Conference
  on Knowledge Discovery and Data Mining}, KDD '18, page 1244–1253, New York,
  NY, USA, 2018. Association for Computing Machinery.

\bibitem{robustness_paper}
David Alvarez~Melis and Tommi Jaakkola.
\newblock Towards robust interpretability with self-explaining neural networks.
\newblock In S.~Bengio, H.~Wallach, H.~Larochelle, K.~Grauman, N.~Cesa-Bianchi,
  and R.~Garnett, editors, {\em Advances in Neural Information Processing
  Systems}, volume~31. Curran Associates, Inc., 2018.

\bibitem{Yanou}
Yanou Ramon, David Martens, Foster~J. Provost, and Theodoros Evgeniou.
\newblock Counterfactual explanation algorithms for behavioral and textual
  data.
\newblock {\em CoRR}, abs/1912.01819, 2019.

\bibitem{mothilal2021towards}
Ramaravind~Kommiya Mothilal, Divyat Mahajan, Chenhao Tan, and Amit Sharma.
\newblock Towards unifying feature attribution and counterfactual explanations:
  Different means to the same end.
\newblock In {\em AAAI/ACM Conference on AI, Ethics, and Society (AIES)}, April
  2021.

\bibitem{jesus}
S\'{e}rgio Jesus, Catarina Bel\'{e}m, Vladimir Balayan, Jo\~{a}o Bento, Pedro
  Saleiro, Pedro Bizarro, and Jo\~{a}o Gama.
\newblock How can i choose an explainer? an application-grounded evaluation of
  post-hoc explanations.
\newblock In {\em Proceedings of the 2021 ACM Conference on Fairness,
  Accountability, and Transparency}, FAccT '21, page 805–815, New York, NY,
  USA, 2021. Association for Computing Machinery.

\bibitem{many_faces}
Vivian {Lai}, Jon~Z. {Cai}, and Chenhao {Tan}.
\newblock {Many Faces of Feature Importance: Comparing Built-in and Post-hoc
  Feature Importance in Text Classification}.
\newblock {\em arXiv e-prints}, page arXiv:1910.08534, October 2019.

\bibitem{Lin2019DoER}
Zhong~Qiu Lin, Mohammad~Javad Shafiee, Stanislav Bochkarev, Michael~St. Jules,
  Xiao~Yu Wang, and Alexander Wong.
\newblock Do explanations reflect decisions? a machine-centric strategy to
  quantify the performance of explainability algorithms.
\newblock {\em ArXiv}, abs/1910.07387, 2019.

\bibitem{interp_health}
Radwa El~Shawi, Youssef Sherif, Mouaz Al-Mallah, and Sherif Sakr.
\newblock Interpretability in healthcare a comparative study of local machine
  learning interpretability techniques.
\newblock In {\em 2019 IEEE 32nd International Symposium on Computer-Based
  Medical Systems (CBMS)}, pages 275--280, 2019.

\bibitem{more_health}
Jamie Duell, Xiuyi Fan, Bruce Burnett, Gert Aarts, and Shang-Ming Zhou.
\newblock A comparison of explanations given by explainable artificial
  intelligence methods on analysing electronic health records.
\newblock In {\em 2021 IEEE EMBS International Conference on Biomedical and
  Health Informatics (BHI)}, pages 1--4, 2021.

\bibitem{fooling_lime_shap}
Dylan Slack, Sophie Hilgard, Emily Jia, Sameer Singh, and Himabindu Lakkaraju.
\newblock Fooling lime and shap: Adversarial attacks on post hoc explanation
  methods.
\newblock In {\em AAAI/ACM Conference on Artificial Intelligence, Ethics, and
  Society (AIES)}, 2020.

\bibitem{domen_robust}
Domen Vres and Marko Robnik{-}Sikonja.
\newblock Better sampling in explanation methods can prevent dieselgate-like
  deception.
\newblock {\em CoRR}, abs/2101.11702, 2021.

\bibitem{limitations}
Christoph Molnar.
\newblock {\em Limitations of Interpretable Machine Learning Methods}.
\newblock 2020.
\newblock \url{https://github.com/compstat-lmu/iml_methods_limitations}.

\bibitem{bishop_bias_var}
Christophper~M. Bishop.
\newblock {\em Pattern Recognition and Machine Learning}.
\newblock 2006.
\newblock Chapter 3, Section 2.

\bibitem{multi-vae}
Dawen Liang, Rahul~G. Krishnan, Matthew~D. Hoffman, and Tony Jebara.
\newblock Variational autoencoders for collaborative filtering.
\newblock In {\em Proceedings of the 2018 World Wide Web Conference}, WWW '18,
  page 689–698, Republic and Canton of Geneva, CHE, 2018. International World
  Wide Web Conferences Steering Committee.

\bibitem{movielens}
F.~Maxwell Harper and Joseph~A. Konstan.
\newblock The movielens datasets: History and context.
\newblock {\em ACM Trans. Interact. Intell. Syst.}, 5(4), December 2015.

\bibitem{Sippy2020DataSA}
Jacob Sippy, Gagan Bansal, and Daniel~S. Weld.
\newblock Data staining: A method for comparing faithfulness of explainers.
\newblock 2020.

\bibitem{cxplain}
Patrick Schwab and Walter Karlen.
\newblock Cxplain: Causal explanations for model interpretation under
  uncertainty.
\newblock In H.~Wallach, H.~Larochelle, A.~Beygelzimer, F.~d\textquotesingle
  Alch\'{e}-Buc, E.~Fox, and R.~Garnett, editors, {\em Advances in Neural
  Information Processing Systems}, volume~32. Curran Associates, Inc., 2019.

\end{thebibliography}

%\begin{thebibliography}{1}

%\bibitem{kour2014real}
%George Kour and Raid Saabne.
%\newblock Real-time segmentation of on-line handwritten arabic script.
%\newblock In {\em Frontiers in Handwriting Recognition (ICFHR), 2014 14th
%  International Conference on}, pages 417--422. IEEE, 2014.

%\bibitem{kour2014fast}
%George Kour and Raid Saabne.
%\newblock Fast classification of handwritten on-line arabic characters.
%\newblock In {\em Soft Computing and Pattern Recognition (SoCPaR), 2014 6th
%  International Conference of}, pages 312--318. %IEEE, 2014.

%\bibitem{hadash2018estimate}
%Guy Hadash, Einat Kermany, Boaz Carmeli, Ofer Lavi, George Kour, and Alon
%  Jacovi.
%\newblock Estimate and replace: A novel approach to integrating deep neural
%  networks with existing applications.
%\newblock {\em arXiv preprint arXiv:1804.09028}, %2018.
%
%\end{thebibliography}

\end{document}